\documentclass{article}

\usepackage{PRIMEarxiv}

\usepackage[utf8]{inputenc} 
\usepackage[T1]{fontenc}    
\usepackage{hyperref}       
\usepackage{url}            
\usepackage{booktabs}       
\usepackage{amsfonts}       
\usepackage{nicefrac}       
\usepackage{microtype}      
\usepackage{fancyhdr}       
\usepackage{algorithm}
\usepackage{algpseudocode}
\usepackage{makecell,multirow,threeparttable}
\usepackage{amsmath}
\usepackage{adjustbox}
\usepackage{subcaption}
\usepackage{multicol}
\usepackage{orcidlink}
\usepackage{graphicx}       
\graphicspath{{figs/}}     

\algblock{Input}{EndInput}
\algnotext{EndInput}
\algblock{Output}{EndOutput}
\algnotext{EndOutput}
\newcommand{\Desc}[2]{\State \makebox[2em][l]{#1}#2}

\title{Towards Superior Quantization Accuracy: A Layer-sensitive Approach
}

\author{
  Feng Zhang\,\orcidlink{0009-0007-1891-6403},
  Yanbin Liu\,\orcidlink{0000-0003-4724-8065},
  Weihua Li\,\orcidlink{0000-0001-9215-4979},
  Jie Lv\,\orcidlink{0009-0001-8713-3660},
  Xiaodan Wang\,\orcidlink{0009-0008-2159-2339},
  Quan Bai\,\orcidlink{0000-0003-1214-6317}
}

\begin{document}
\maketitle

\begin{abstract}

\sloppy Large Vision and Language Models have exhibited remarkable human-like
intelligence in tasks such as natural language comprehension, problem-solving,
logical reasoning, and knowledge retrieval. However, training and serving these
models require substantial computational resources, posing a significant
barrier to their widespread application and further research. To mitigate this
challenge, various model compression techniques have been developed to reduce
computational requirements. Nevertheless, existing methods often employ uniform
quantization configurations, failing to account for the varying difficulties
across different layers in quantizing large neural network models. This paper
tackles this issue by leveraging layer-sensitivity features, such as activation
sensitivity and weight distribution Kurtosis, to identify layers that are
challenging to quantize accurately and allocate additional memory budget. The
proposed methods, named SensiBoost and KurtBoost, respectively, demonstrate
notable improvement in quantization accuracy, achieving up to 9\% lower
perplexity with only a 2\% increase in memory budget on LLama models compared
to the baseline.

\end{abstract}


\section{Introduction}


Large Language Models (LLMs) have significantly advanced artificial
intelligence, demonstrating human-like capabilities in natural language
comprehension, problem-solving, logical reasoning, and knowledge retrieval.
These models power a wide range of applications, from chatbots and virtual
assistants to code generation and scientific discovery. However, their
deployment is hindered by substantial computational and memory demands, which
necessitates the needs for efficient model compression and quantization
techniques to low the bar of entry.

Quantization techniques aim to reduce the memory footprint and computational
requirements of LLMs while preserving their performance. Existing quantization
methods, such as AWQ~\cite{lin_awq_2023}, GPTQ~\cite{frantar_gptq_2023},
BnB~\cite{dettmers_qlora_2023}, and HQQ~\cite{badri2023hqq}, predominantly
employ uniform quantization configurations. While effective to some extent,
these approaches fail to consider the varying quantization difficulty across
different layers of billion-scale models.


\begin{figure*}[t]
  \centering
\begin{subfigure}[t]{0.48\textwidth}
\includegraphics[width=\textwidth]{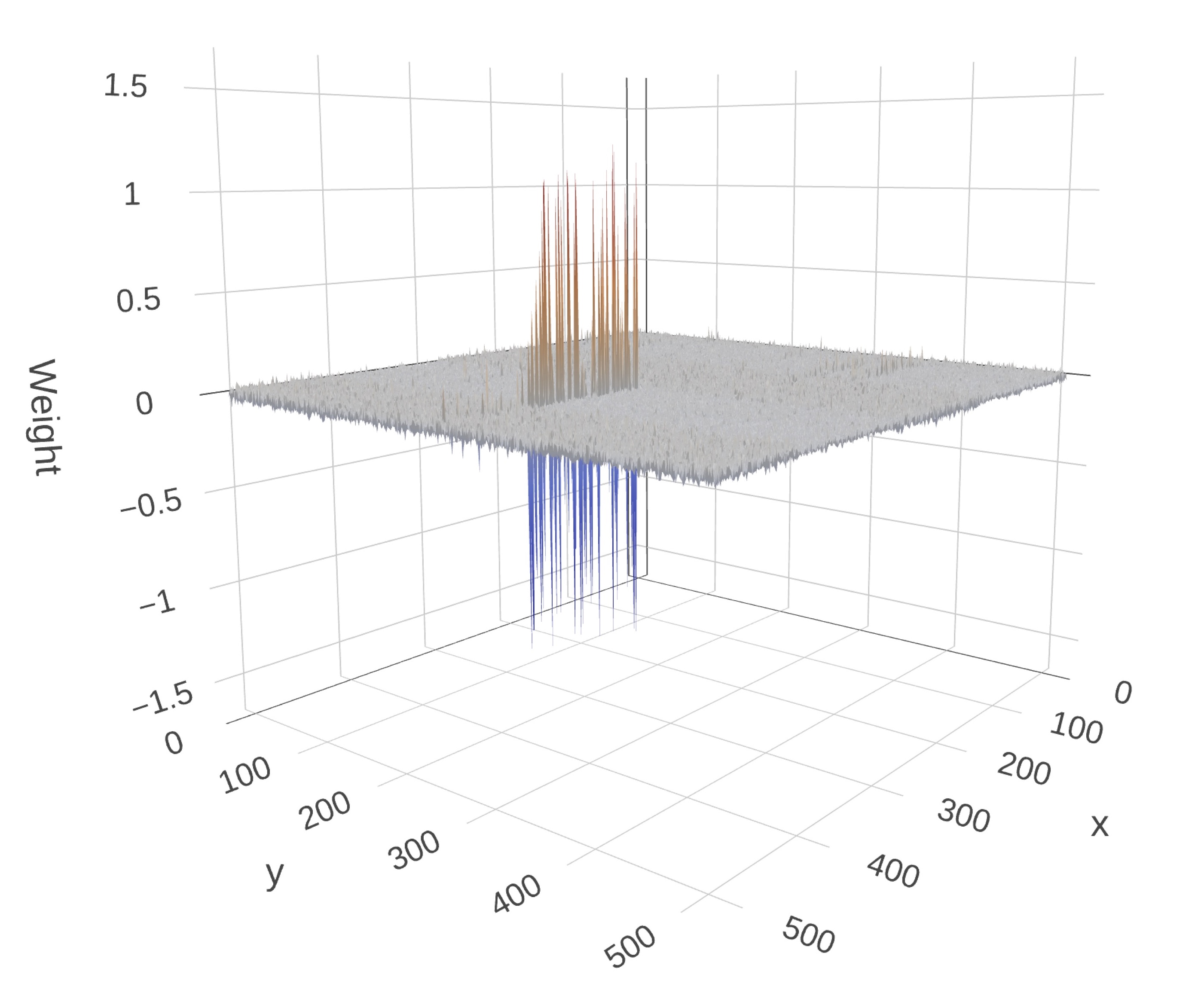}
\caption[3D illustration of weights in last layer]{
This figure demonstrates the subset of the weights (the $512 \times 512$
sub-region centered at $(2533, 3037)$) in the second layer of self-attention output
projection module of the Llama-2-7B model. The presence of extensive long
spikes indicates significant outliers inside the layer of the self attention
output projection module.
}
\label{fig:llama-7b-o-proj-32}
\end{subfigure}
~
\begin{subfigure}[t]{0.48\textwidth}
\includegraphics[width=\textwidth]{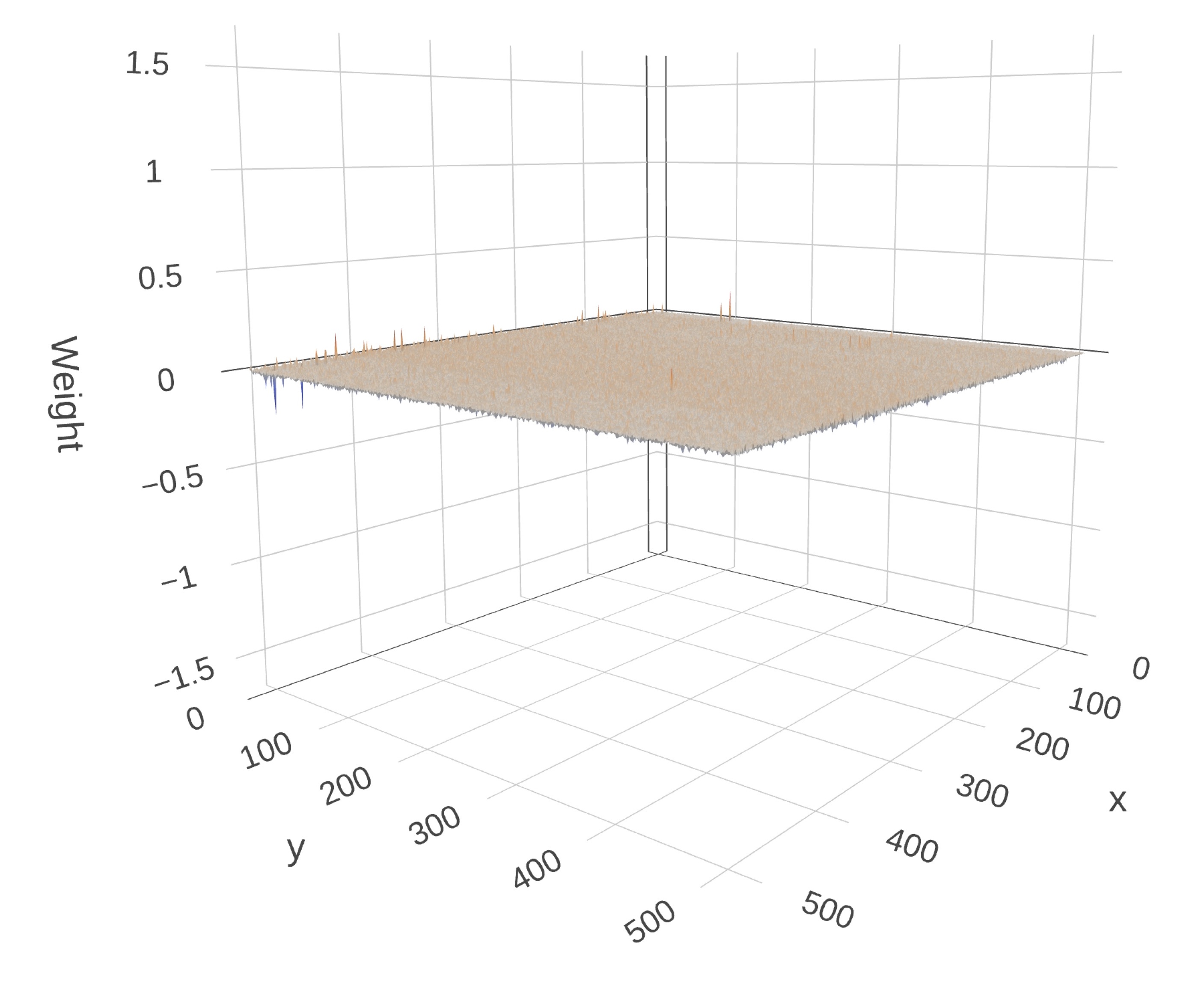}
\caption[3D illustration of weights in second layer]{
This figure shows a flat plane of subset of the weights (the $512 \times 512$
sub-region centered at $(2533, 3037)$) in the second layer of self-attention output
projection module of the Llama-2-7B model.
}
\label{fig:llama-7b-o-proj-2}
\end{subfigure}

\end{figure*}

Deep neural network model's weights are typically initialized using the
Kaiming~\cite{he_delving_2015} or the Xavier initialization
method~\cite{glorot_understanding_2010}, leading to a zero-centered normal
distribution with a standard deviation usually less than 1. Outliers are
introduced during model training due to several reasons. For example, An et
al.~\cite{an2025systematic} revealed that softmax attention is the root cause
of outliers in transformer-based neural network models. Additionally, multiple
studies\cite{kovaleva2021bert,wei2022outlier,elhage2023privileged} discovered
that layer normalization contributes to the introduction of outliers. Weights
with significant outliers are challenging to quantize accurately since the
accommodation of outliers squeezes most normal weights into a narrower range,
resulting in an imprecise representation of these weights. The uneven presence of
outliers across layers leads to varying quantization difficulty across
layers in a particular LLM. As illustrated in
Figure~\ref{fig:llama-7b-o-proj-32} and Figure~\ref{fig:llama-7b-o-proj-2}, the
weight magnitudes in the \verb=self_attn.o_proj= module differ significantly
between the second layer and the last layer. While the last layer shows
substantial outliers, the second layer exhibits no notable outliers. This
suggests that a uniform quantization approach may not be optimal.

To address this, MXQ~\cite{zhang2025mxq} introduced a mixed-integer linear
programming (MiLP) based approach to assign differentiated quantization
configurations while maintaining an overall memory budget. However, despite its
adaptive allocation strategy, MXQ-quantized models often underperform compared
to the baseline methods such as HQQ, BnB, and AWQ, indicating that the MXQ
quantization approach may not effectively prioritize quantization accuracy over
memory efficiency in the trade-off.

Additionally, prior research~\cite{samragh2023weight, xu2023initializing,
molchanov2019importance} has shown that the importance of weights within a deep
neural network is non-uniform. Motivated by these observations and the
limitations of the state-of-the-art quantization techniques, this paper
introduces a novel approach based on layer sensitivity analysis. We hypothesize
that memory allocation contributes equally to quantization accuracy across most
layers in LLMs, but a subset of layers, termed sensitive layers, require
additional memory to maintain optimal performance. Identifying these layers and
selectively allocating extra memory resources can enhance overall quantization
accuracy with minimal additional cost.

Our proposed method leverages layer-wise sensitivity metrics, including
activation sensitivity (hereafter referred to as "sensitivity") and weight
distribution kurtosis~\cite{decarlo_meaning_1997}, to identify demanding
layers. By selectively allocating additional memory to these layers while
slightly relaxing the overall memory constraint, we achieve improved
quantization accuracy without incurring significant overhead. Our main
contributions are as follows:
\begin{itemize}
\item We empirically explored layer-wise activation sensitivity to quantization
  error on multiple transformer-based LLM families, revealing the robustness of
  sensitivity within a family of models and their fine-tuned variants.
\item We proposed a simple outlier detection algorithm to discover sensitive
  layers with activation sensitivity scores or Kurtosis metrics.
\item Based on the outlier detection algorithm, we proposed the SensiBoost and
  KurtBoost methods that outperform HQQ with a reduction in perplexity up to
  9\% while increasing memory budget only by 2\%.
\end{itemize}

\section{Related Works}



\begin{figure*}[t]
  \centering
\begin{subfigure}[t]{0.48\textwidth}
\includegraphics[width=\textwidth]{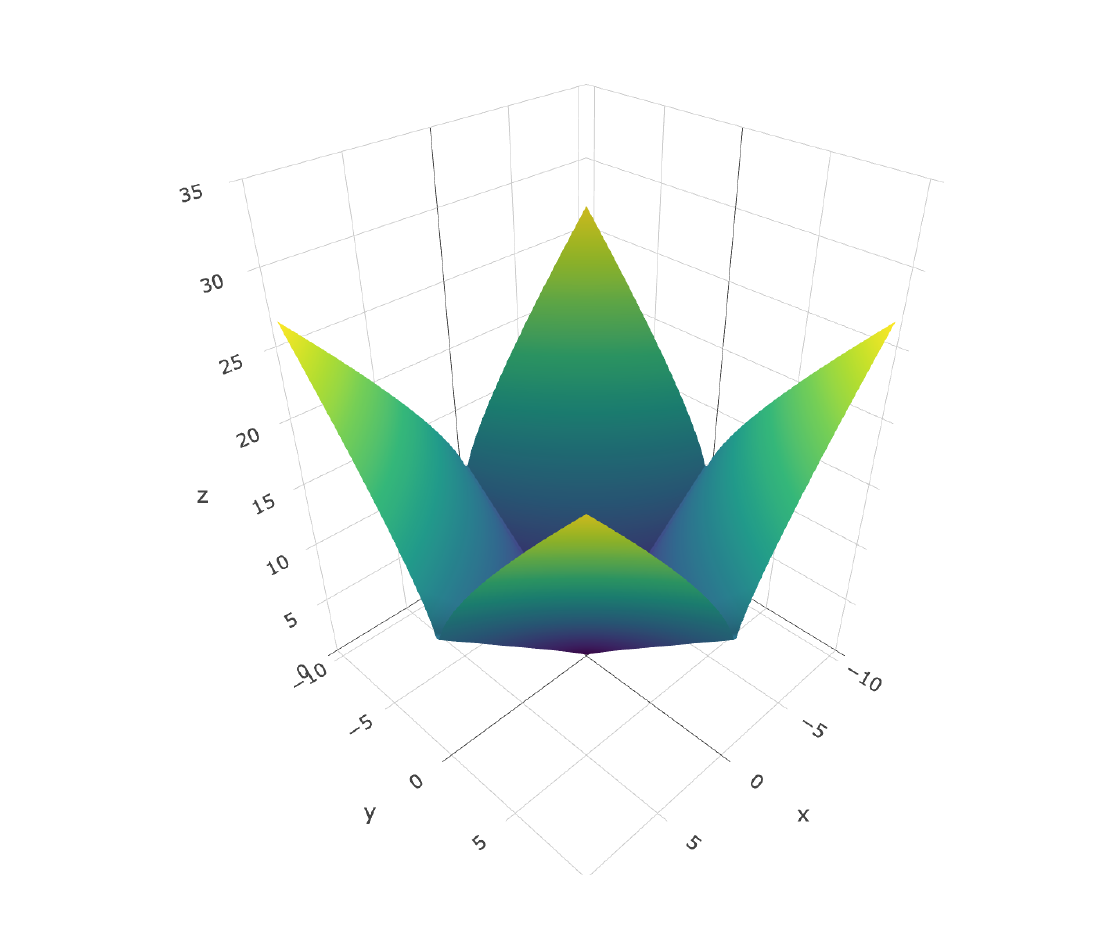}
\caption[$L_{p=0.7}$-norm 3D visualization]{ This figure illustrates the
visualization of the two-variable $L_{p=0.7}$-norm function as a surface in 3D
space. This $L_p$-norm is employed by HQQ to preserve outliers in weights of LLMs.}
\label{fig:lpnorm-07-visual}
\end{subfigure}
~
\begin{subfigure}[t]{0.48\textwidth}
\includegraphics[width=\textwidth]{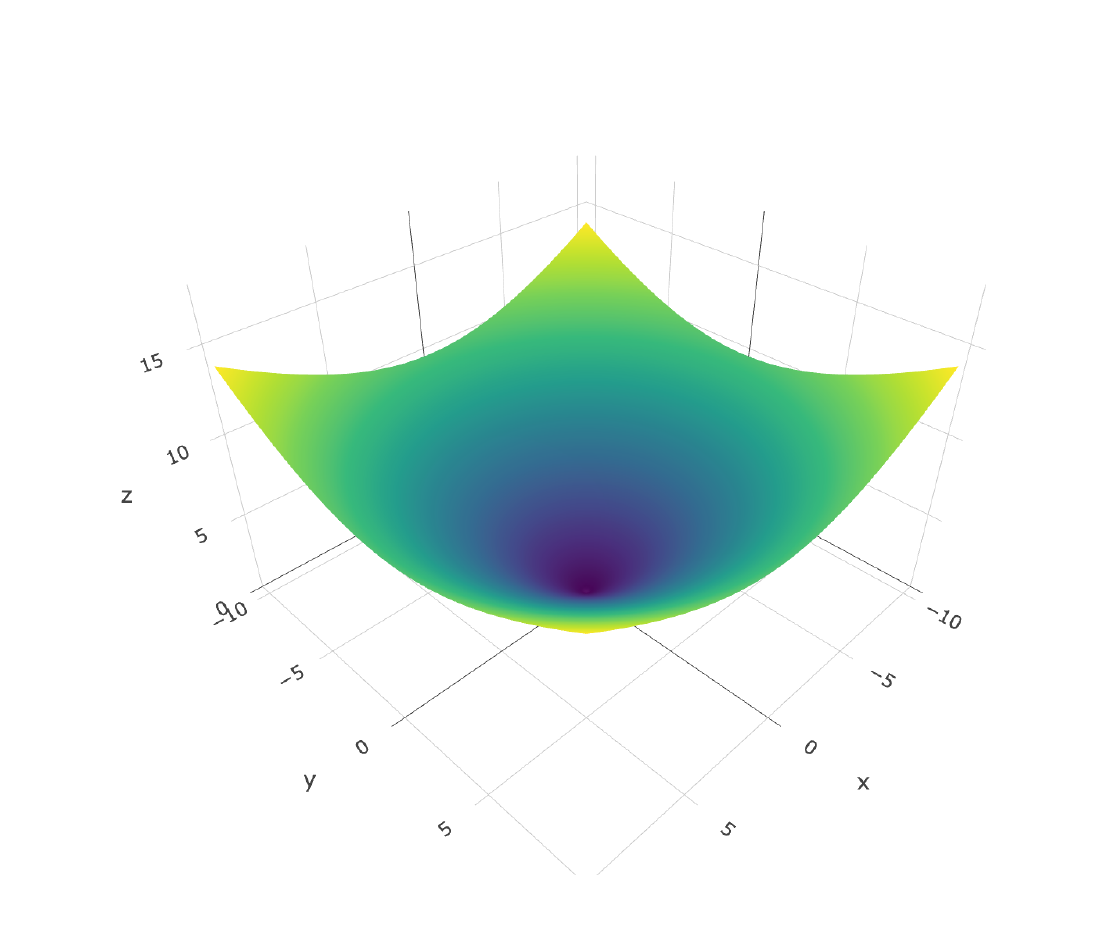}
\caption[$L_{p=2}$-norm 3D visualization]{ This figure illustrates the
visualization of the two-variable $L_{p=2}$-norm function as a surface in 3D
space.}
\label{fig:lpnorm-squred-visual}
\end{subfigure}\end{figure*}

\sloppy The quantization techniques have been extensively studied to address a
wide range of use cases, including
inference~\cite{frantar_gptq_2023,lin_awq_2023}, KV cache
compression~\cite{hooper2024kvquant},
fine-tuning~\cite{dettmers_qlora_2023,guo_lq-lora_2024} and optimizer state
\cite{dettmers_8-bit_2022}. These techniques can be generally divided into
two categories: \textbf{1)} Quantization Aware Training (QAT)
\cite{nagel_white_2021}, which is tightly coupled with the resource-intensive
and time-consuming training process, and \textbf{2)} Post Training
Quantization (PTQ) \cite{nagel_data-free_2019}, a training-free method. In this
paper, we focus on a specific class of the PTQ method known as weight-only
quantization method. Specifically, weight-only quantization can be further
categorized into calibration-based and calibration-free methods, depending on
whether an additional calibration dataset is adopted during quantization. The
following section discusses the weight-only quantization methods.

\subsection{Calibration-based Methods}

The calibration-based approaches leverage the Hessian matrix and Fisher
information. While often achieving better quantization accuracy, they tend to
be slow and challenging to generalize to models with distinct architectures.
The representative state-of-the-art implementations of calibration-based
approaches include GPTQ and AWQ.

\noindent\textbf{GPTQ} \cite{frantar_gptq_2023} is based on Optimal Brain
Quantizer (OBQ)~\cite{frantar_obq_2022}, which quantized one weight at a time
while constantly updating all not-yet-quantized weights to compensate for the
error incurred by quantizing a single weight. GPTQ improves OBQ by quantizing
weight column-wise to eliminate repeated calculation of the inverse of the
Hessian Matrix, thus scaling to larger models with parameters as large as a few
hundreds of billions. GPTQ has extensively optimized kernels to accelerate
mixed-precision matrix multiplication. Thus, the GPTQ quantized models not only
save memory but also run faster.

\noindent\textbf{AWQ} \cite{lin_awq_2023} proposes a quantization method to
identify the small fraction of ``salient'' weights by measuring activation
magnitude and pre-scaling the weights with a per-channel factor $s$ to minimize
quantization errors based on the observation that the significance of LLM's
weights is non-uniform. Equation~\ref{eq:awq-obj} formulates the objective
function of AWQ:
\begin{equation}
  \begin{aligned}
    s^* &= \underset{s}{argmin}\, \mathcal{L}(s) \\
    \mathcal{L}(s) &= \|Q(\mathrm{W}\cdot s)(s^{-1}\cdot \mathrm{X}) - \mathrm{W}\mathrm{X}\| \\
    Q(\mathrm{W}) &= \Delta \cdot Round(\frac{\mathrm{W}}{\Delta}) \\
    \Delta &= \frac{max(|\mathrm{W}|)}{2^{n-1}} \\
  \end{aligned}
  \label{eq:awq-obj}
\end{equation}

To tackle the non-differentiability of the loss function in Equation~\ref{eq:awq-obj}, AWQ
leverages a simple search space, where $\alpha$ is confined to the interval $[0,
1]$, as defined in Equation~\ref{eq:awq-search-space} to find the optimal scale
$s$.

\begin{equation}
  s = sx^{\alpha}, \qquad \alpha^* = \underset{\alpha}{argmin}\, \mathcal{L}(sx^{\alpha})
  \label{eq:awq-search-space}
\end{equation}

This approach unifies the treatment of the salient and non-salient weights,
eliminating the need to isolate salient weights into separate storage like
sparse matrix, and develop specialised mixed-precision matrix multiplication
kernel for fast inference. Besides enabling significant memory reduction, AWQ
achieves approximately 3 times inference acceleration compared to the FP16
implementation by Huggingface across a wide range of LLMs.

\subsection{Calibration-free Methods}

\noindent\textbf{HQQ} \cite{badri2023hqq} leverages quantization parameters
zero-point $z$ and scaling $s$ to minimize the $L_{p<1}$-norm between the
original weights $\mathrm{W}$ and their dequantized counterpart as defined in
Equation~\ref{eq:hqq-objective1}.

\begin{equation}
  \underset{z,s}{argmin}\phi(\mathrm{W} - Q_{z,s}^{-1}(Q_{z,s}(\mathrm{W})))
  \label{eq:hqq-objective1}
\end{equation}

The incorporation of the $L_{p<1}$-norm in the loss function $\phi()$ enables
HQQ to model outliers effectively through a hyper-Laplacian distribution, which
captures the long-tailed nature of outliers more accurately than the
conventional squared error. Figure~\ref{fig:lpnorm-07-visual} illustrates the
non-convex nature of $L_{p=0.7}$-norm (employed by HQQ) as a 3D surface.
Figure~\ref{fig:lpnorm-squred-visual} shows the 3D plot of the $L_{p=2}$-norm,
a convex funtion. The $L_{p<1}$-norm makes the loss function $\phi()$
non-convex. Therefore, HQQ converts the optimization of the non-convex loss
function $\phi()$ formulated in Equation~\ref{eq:hqq-objective1} to a new
formulation denoted in Equation~\ref{eq:hqq-hq-obj} so that it can leverage the
Half-Quadratic solver \cite{geman1992conres}.

\begin{equation}
  \underset{z,\mathrm{W}_e}{argmin}\phi(\mathrm{W}_e) + \frac{\beta}{2}\Big\|\mathrm{W}_e - (\mathrm{W} - Q_{z}^{-1}(Q_{z}(\mathrm{W})))\Big\|_2^2
  \label{eq:hqq-hq-obj}
\end{equation}

By utilizing alternate optimization, Equation~\ref{eq:hqq-hq-obj} is
decomposed into two sub-problems as illustrated in Equation~\ref{eq:hqq-hq-sp}.

\begin{equation}
  \begin{alignedat}{3}
    \mathrm{W}_e^{(t+1)} & \leftarrow \underset{\mathrm{W}_e}{argmin}\phi(\mathrm{W}_e) + \frac{\beta^{(t)}}{2}\Big\|\mathrm{W}_e - (\mathrm{W} - Q_{z}^{-1}(Q_{z}(\mathrm{W})))\Big\|_2^2 & \, (sp_1) \\
    z^{(t+1)} & \leftarrow \underset{z}{argmin}\frac{1}{2}\Big\|Q_{z}^{-1}(Q_{z}(\mathrm{W})) - (\mathrm{W} - \mathrm{W}_e^{(t+1)})\Big\|_2^2 & \, (sp_2)  \\
    \beta^{(t+1)} & \leftarrow k\beta^{(t)} & \\
  \end{alignedat}
  \label{eq:hqq-hq-sp}
\end{equation}

When $L_{p<1}$-norm is the loss function, the solution to the first
sub-problem($sp_1$) is the generalized soft-thresholding
operator\cite{badri2016non} as illustrated in
Equation~\ref{eq:hqq-soft-threshold}.

\begin{equation}
  \begin{alignedat}{2}
    \mathrm{W}_e^{(t+1)} \quad & \leftarrow \quad shrink_{l_p} (\mathrm{W} - Q_{z}^{-1}(Q_{z}(\mathrm{W})), \beta) \\
    shrink_{l_p}(x, \beta) \quad & = \quad sign(x) relu\Big(|x| - \frac{|x|^{p-1}}{\beta}\Big) \\
  \end{alignedat}
  \label{eq:hqq-soft-threshold}
\end{equation}

The second sub-problem($sp_2$) can be converted to Equation~\ref{eq:hqq-sp2}.
The solution, as presented in Equation~\ref{eq:hqq-sp2-sol}, is the average
over the axis where the quantization grouping is carried out.

\begin{equation}
  \begin{alignedat}{2}
  z^{(t+1)} \quad & \leftarrow \quad \underset{z}{argmin}\frac{1}{2}\Big\|z-(\mathrm{W}_q^{(t+1)-\frac{\mathrm{W}-\mathrm{W}_e^{(t+1)}}{s}})\Big\|_2^2 \\
  \mathrm{W}_q^{(t+1)} \quad & = \quad round(\mathrm{W}/s+z^{(t)})\\
  \end{alignedat}
  \label{eq:hqq-sp2}
\end{equation}

\begin{equation}
  \begin{aligned}
    z^{(t+1)} \quad & \leftarrow & \Big\langle \mathrm{W}_q^{(t+1)}-\frac{\mathrm{W}-\mathrm{W}_e^{(t+1)}}{s} \Big\rangle \\
  \end{aligned}
  \label{eq:hqq-sp2-sol}
\end{equation}

HQQ's sole reliance on the weight without considering the layer activation
enables it to generalize to models with diverse underlying architectures. HQQ
achieves comparable performance compared to the top quantization methods such
as AWQ, GPTQ, and BnB while exhibiting extraordinary quantization speed.
Experiments show that HQQ is approximately an order of magnitude faster than
the state-of-the-art calibration-based methods such as AWQ and GPTQ. In
addition, HQQ offers an abundance of options to further optimize the
quantization with a wide range of bits to quantize large neural network models.
Available bit choices include 2, 3, 4 and 8. It also allows configurable group
size, secondary bit and group size for metadata quantization. Furthermore, by
adopting the calibration-free approach, HQQ avoids potential over-fitting to
calibration datasets, making it model architecture-agnostic and generalizable
to not only diverse transformer-based large language models but also
multi-modal models.

\noindent\textbf{BnB} \cite{dettmers_qlora_2023} employs a novel high-precision
technique to quantize pre-trained model weights to 4-bit NormalFloat (NF4),
which employs the Gaussian distribution exhibited in model weights. The 4-bit
NormalFloat datatype represents 16 values ($q1, q2, \cdots, q16$) in the
interval $[-1, 1]$. Each weight matrix is chunked into small groups for better
quantization accuracy. Additionally, NF4 employs the double quantization
technique to reduce the overhead introduced by the granular grouping scheme, a
widely adopted strategy by other state-of-the-art quantization methods.

\noindent\textbf{MXQ} \cite{dettmers_qlora_2023} allocates optimal
configurations that minimize the sum of Frobenius norm of the difference
between the full-precision weight matrices and their quantized counterparts
while maintaining the overall memory consumption within constraints set by a
global bit budget per parameter. MXQ can be formulated as a Mixed-Integer
Linear Programming~\cite{huangfu_parallelizing_2018} problem as denoted in
Equation~\ref{eq:objective-function}:
\begin{equation}
  \begin{aligned}
    \underset{c_1, c_2, \cdots, c_N }{\arg\min} & \sum_{ \substack{i \in \{1, \dots N\} \\ {c_i \in C} }}
    \left\|W^{(i)}-\hat{W}^{(i)}_{c_i}\right\|_{F} \\
    \mathrm{s.t.\qquad}  & \sum_{\substack{i \in \{1, \dots N\} \\ {c_i \in C} }} \texttt{stor}(W^{(i)}, c_i) \le \beta, \\
    \mathrm{where\qquad}  & \texttt{stor}(W^{(i)}, c_i) = |W^{(i)}| \cdot \left(b_1 + \frac{2b_2}{g1} + \frac{32}{g_1 \cdot g_2}\right) \\
  \end{aligned}
  \label{eq:objective-function}
\end{equation}

where $c_i = (b_1, g_1, b_2, g_2)$ denotes the configuration parameters used to
quantize the $i$th matrix of the LLM, $b_1$ and $g_1$ represent the bit width
and group size for quantizing weights, $b_2$ and $g_2$ indicate the bit width
and group size to quantize metadata, such as zero points and scales. $C$ is the
set of 12 possible configurations. Additionally, $N$ is the number of weight
matrices to be quantized, $W^{(i)}$ and $\hat{W}^{(i)}$ are the $i$th
full-precision and quantized weight matrices, respectively. The parameter
$\beta$ denotes the overall memory budget in megabytes. By introducing $M=|C|
\times N$ binary decision variables, Equation~\ref{eq:objective-function} is
further converted to standard LP formulation~\cite{boyd2004convex} so that it
can be solved efficiently by off-the-shelf LP solvers such as
Gurobi~\cite{gurobi} and HiGHS~\cite{huangfu_parallelizing_2018}.

Except for the relatively new MXQ approach, these methods have been adopted
extensively in the industry, demonstrating their practicality and efficacy.
Nevertheless, the limitations of these methods are worth discussing. First,
quantization methods such as GPTQ and AWQ require curated calibration datasets,
making it challenging to generalize these methods to other large neural networks
such as vision models, which are trained on a mixture of textual and image data.
Given the substantial architectural disparities and diverse choices of training
datasets for these multi-modal models, curating compatible calibration datasets
is definitely a maintenance headache. Second, calibration-dependent approaches
tend to rely on GPUs to perform the quantization as a full inference pass is
indispensable to measure the quantization error in terms of activation. This
prevents offloading the quantization task to CPUs, which is cheaper and
more accessible. Additionally, the quantization speed of calibration dataset-dependent methods like AWQ and GPTQ is relatively slow. For instance, the GPTQ
method takes approximately 4 GPU hours to quantize the OPT-175B or BLOOM-176B
models~\cite{frantar_gptq_2023}. Finally, the first four quantization methods
surveyed in this chapter employ uniform quantization configurations across the
entire model, which may be sub-optimal to address varying difficulty across
diverse layers of billion-scale LLMs.

\section{Layer-sensitive Quantization}
\subsection{Activation Sensitivity}

Transformer-based large language models are composed of multiple layers or
blocks~\cite{vaswani_attention_2017}. Each layer consists of the self-attention
and Multi-Layer Perceptron (MLP, a.k.a. FFN) sub-layers. Specifically, the
Llama family model's self-attention includes weights for K, Q, V, and O
projections, known as \verb=k_proj=, \verb=q_proj=, \verb=v_proj=, and
\verb=o_proj= respectively. Similarly, the MLP sub-layer is composed of weights
referred to as \verb=mlp_proj=, \verb=mlp_down=, and \verb=mlp_gate=. The
weights in a large language model are not equally important as revealed by the
observation from prior study \cite{lin_awq_2023}, which claims that preserving
a small portion of so-called salient weights can significantly improve the
quantization accuracy. These weights correspond to particular channels inside
a weight matrix. Motivated by this finding, this paper hypothesizes that there
also exist sensitive layers that are more severely affected by weight
perturbation than others. Protecting such layers by allocating a larger bit
budget will result in an improvement in overall quantization accuracy.

\textbf{Activation Sensitivity Score} In this section, we define Activation
Sensitivity Score, formulated in Equation~\ref{eq:sensitivity-eq}, as mean
squared error between the activations obtained by multiplying the original and
quantized weight with the input. This metric quantifies layer-wise sensitivity
to perturbation introduced by quantization error of a particular model.

\begin{equation}
  s_i = \frac{\Big\|W_i\cdot X - Q^{-1}(Q(W_i))\cdot X\Big\|_2^2}{|W_i\cdot X|}
  \label{eq:sensitivity-eq}
\end{equation}

In Equation~\ref{eq:sensitivity-eq}, $W_i$ denotes the weight in the $i$th
layer, $X$ is the input to the model, which is from a small calibration dataset.
The $Q()$ function represents the quantization function to convert the
full-precision weight into its quantized counterpart. The $Q^{-1}()$ function
is the inverse of the $Q()$.


\begin{figure}[t]
  \centering
\includegraphics[width=0.95\textwidth]{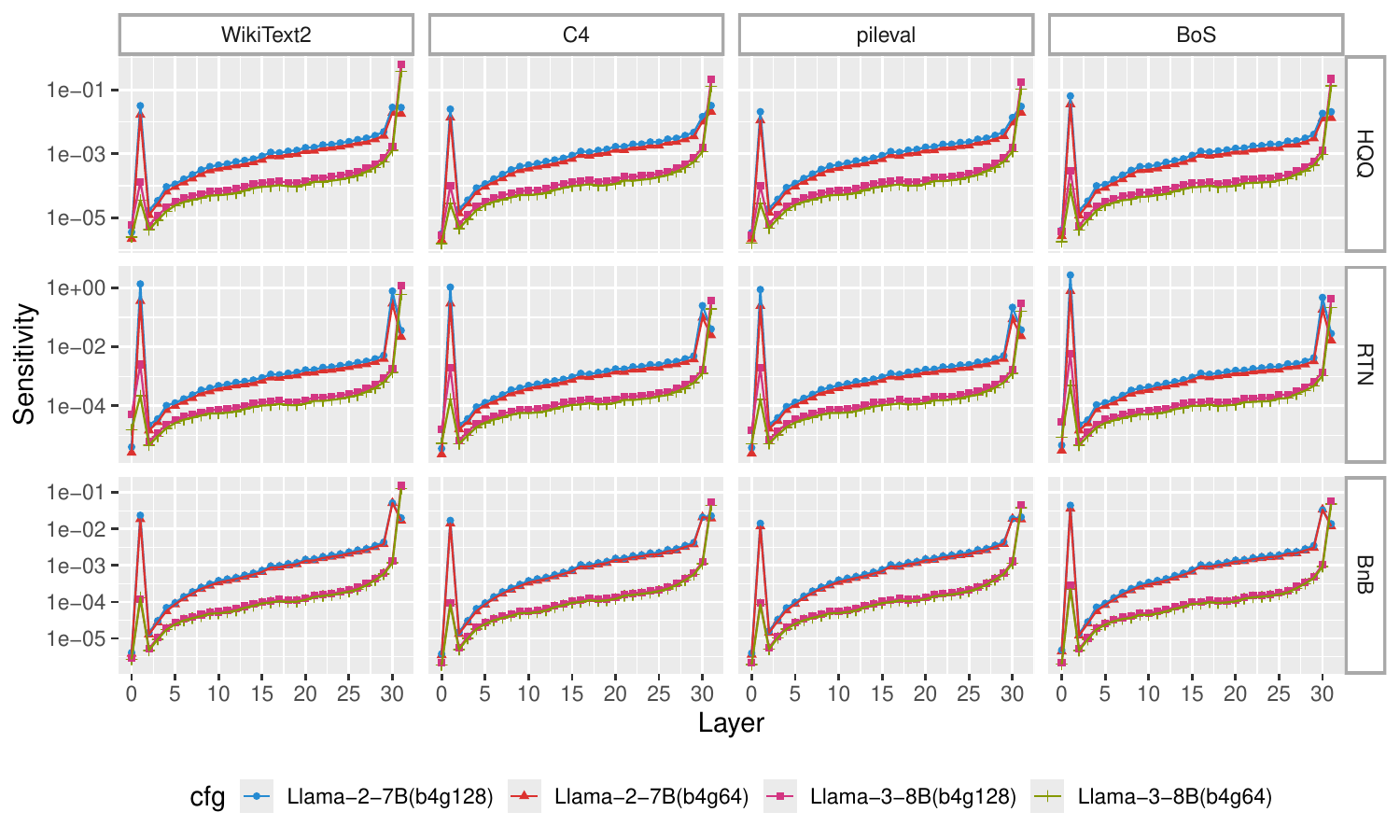}
\caption[Relationship between quantization method, dataset and layer-wise sensitivity]{
  This figure demonstrates the relationship between quantization methods (HQQ,
  RTN, BnB), datasets (WikiText2, C4 pileval, BoS) and layer-wise
  sensitivity. The distinct shapes of sensitivity curves for Llama-2-7B and
  Llama-3-8B models indicate the sensitivity property is model dependent. Meanwhile,
  the near identical patterns across calibration datasets and quantization
  methods show that layer-wise sensitivity to quantization error is independent of
  calibration datasets and quantization methods. For optimal clarity, the figure
  is best viewed in color and with zoom.
}
\label{fig:sensi-ds-meth}
\end{figure}


\begin{figure}[t]
  \centering
\includegraphics[width=0.95\textwidth]{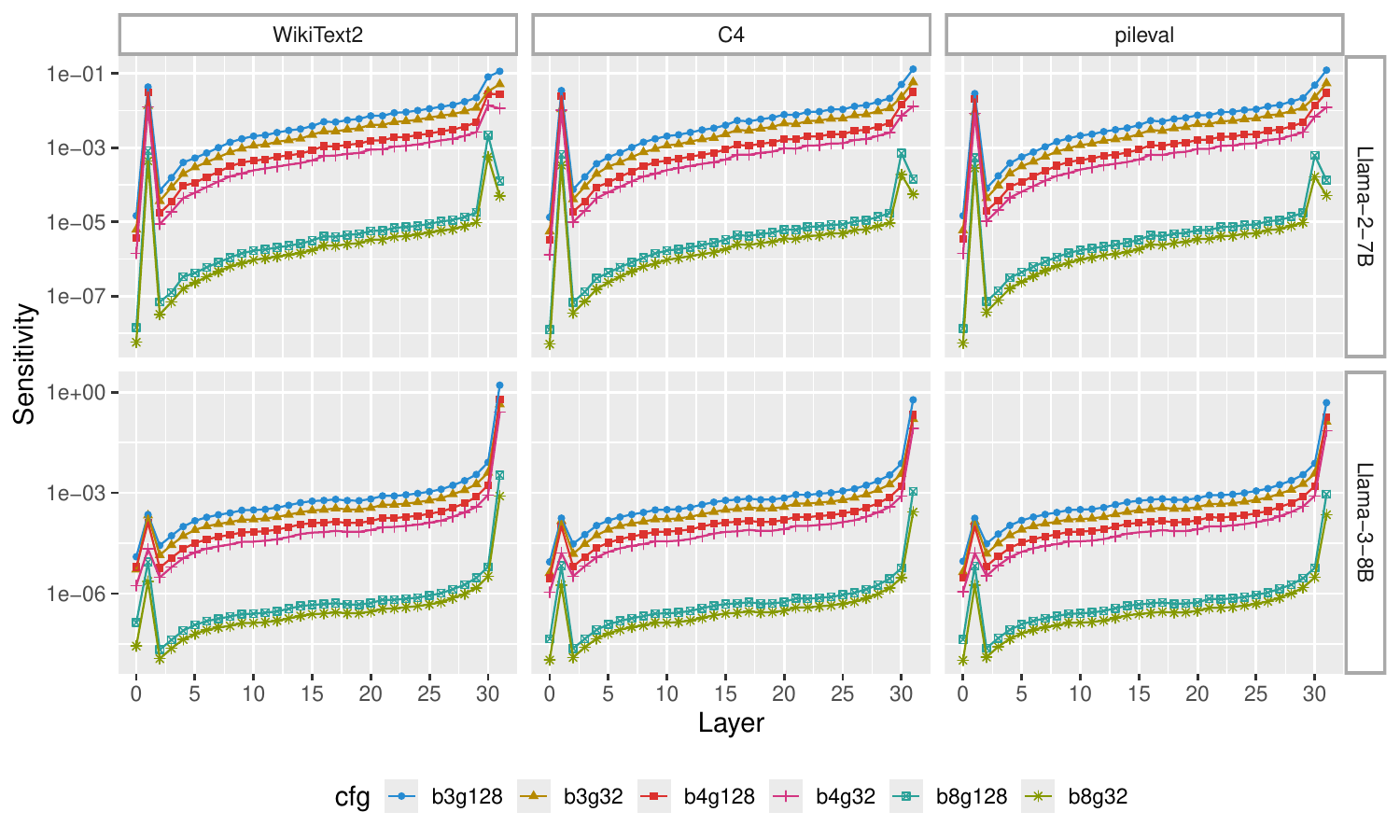}

\caption[Relationship between bit budget layer-wise sensitivity]{
  This figure illustrates how bit budget influnces layer-wise sensitivity. The
  magnitude of sensitivity varies among the 3-bit, 4-bit, and 8-bit groups. The
  4-bit and 8-bit groups show larger difference as indicated by the wider
  blank. However, the overall patterns of the three bit groups demonstrate
  close resemblance. For optimal clarity, the figure is best viewed in color
  and with zoom.
}

\label{fig:sensi-bit-budget}
\end{figure}


\begin{figure}[t]
  \centering
\includegraphics[width=0.95\textwidth]{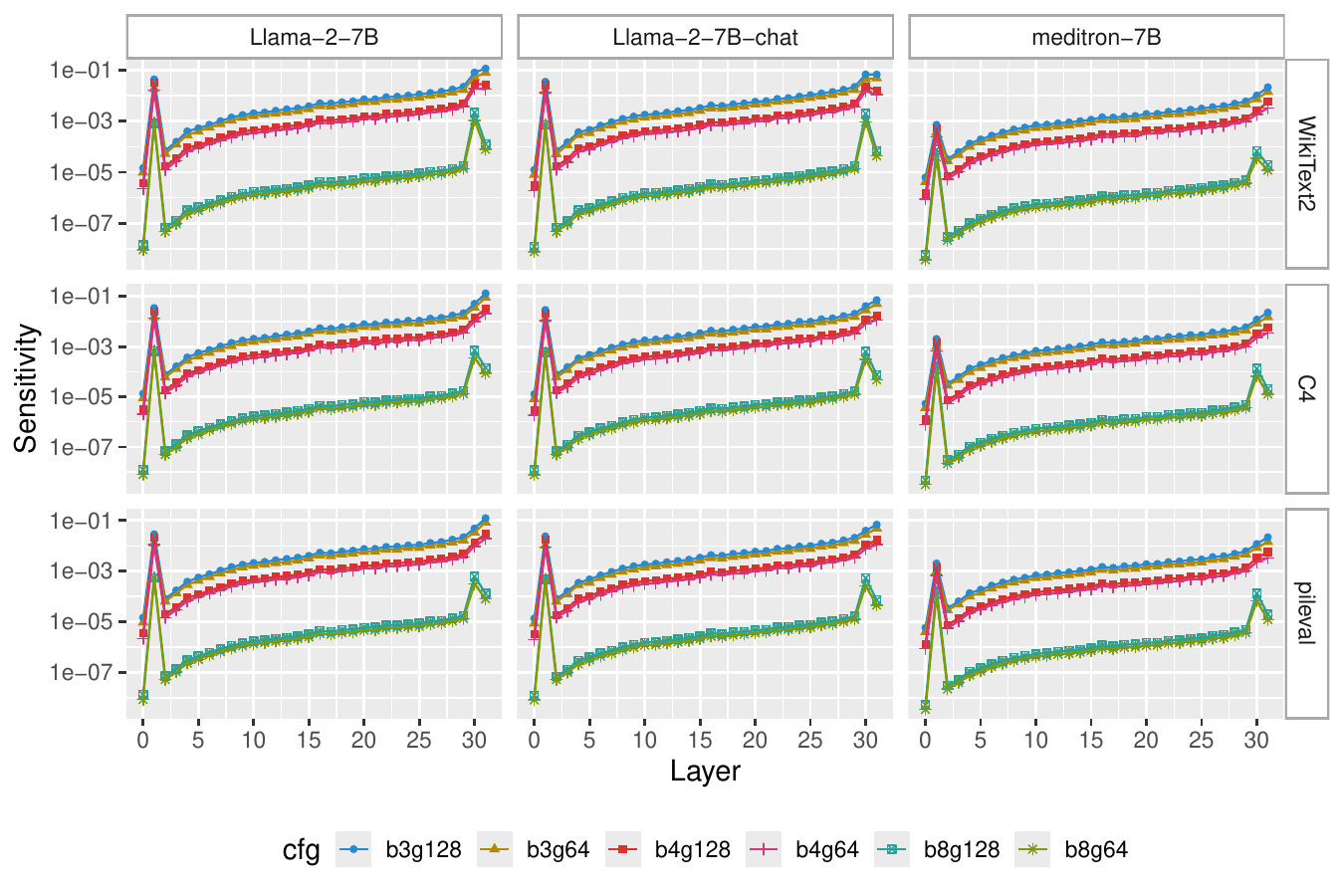}
\caption[Sensitivity inheritance among Llama-2-7B and its fine-tuned variants]{
  This figure presents the sensitivity patterns among the Llama-2-7B base model
  and its fine-tuned mutations. Two fine-tuned models are included for
  comparison. The middle one is Llama-2-7B-chat. And the right is the
  meditron-7B which is a medical LLM fine-tuned on a carefully curated medical
  corpus. As indicated by the nearly identical shapes of sensitivity curves, the
  two fine-tuned models clearly inherit the sensitivity properties from the
  base model. For optimal clarity, the figure is best viewed in color and with
  zoom.
}
\label{fig:sensi-llama27b}
\end{figure}


\textbf{Measuring Sensitivity Score} The measurement of sensitivity requires a
collection of small calibration datasets, which are fed into the target LLMs
layer-by-layer to calculate the output under the full-precision and quantized
weights, respectively. Then the mean squared error (MSE) is computed to quantify
the sensitivity. Specifically, the three open datasets, WikiText-2, C4 and
pileval are utilized to evaluate the robustness of sensitivity. Additionally, a
small synthesized dataset named Branch of Science (BoS) is created to further
validate if the sensitivity property generalizes to diverse datasets. The BoS is
a synthesized dataset composed of a few hundred textual definitions for
science, art and business topics such as Mathematics, Physics, Chemistry, Law,
Music and Journalism, among others. It is generated using the Llama-2-7B model. The
details of the program to produce the BoS dataset are described in Appendix
\ref{sec:bos-gen-tool}. The BoS dataset is published on Hugging Face under the
name
\href{https://huggingface.co/datasets/schnell18/branch-of-science}{schnell18/branch-of-science}.
The program to measure sensitivity is adapted from the AutoAWQ project
on GitHub. For brevity, it is explained in Appendix~\ref{sec:sensi-measuring-tool}.

\textbf{Sensitivity Properties} The layer-wise sensitivity demonstrates
considerable robustness according to the diverse experiments we conducted. We
observed that sensitivity is independent of datasets and quantization methods,
as evidenced by Figure~\ref{fig:sensi-ds-meth}. Moreover, the bit budget only
influences the magnitude of sensitivity, not the overall patterns, which
remain approximately identical across distinct bit budgets, as presented in
Figure~\ref{fig:sensi-bit-budget}, where the 3-bit, 4-bit and 8-bit groups
share almost the same spikes in layers at the beginning although they are separated
by a notable blank. Additionally, fine-tuned models preserve the sensitivity of
the base model, which is demonstrated in Figure~\ref{fig:sensi-llama27b}. Finally,
experiments on the Llama family model reveal that sensitivity spikes tend to be
present at the start and end layers. In summary, the sensitivity properties of
large language models can be described as follows:
\begin{enumerate}
\item Sensitivity is independent of the dataset and quantization method.
\item Sensitivity pattern is consistent among distinct bit budgets.
\item Fine-tuned models preserve the sensitivity of the base model.
\item Sensitivity spikes tend to be present at the start and end layers.
\end{enumerate}

\subsection{Kurtosis}
The Kurtosis measures the deviation from the normal distribution in terms of
tailedness and peakedness~\cite{decarlo_meaning_1997}. It can be formulated as
the standardized fourth population moment about the mean (as denoted in
Equation~\ref{eq:kurtosis-sample}).
\begin{equation}
  k = \frac{\sum_{i=1}^{n}(w_i-\bar{W})^4/n}{(\sum_{i=1}^{n}(w_i-\bar{W})^2/n)^2}
  \label{eq:kurtosis-sample}
\end{equation}
where $w_i$ is the $ith$ weights, $n$ is the number of weights and $\bar{W}$ is
the mean of the weights. There are three distinct Kurtosis types:
\begin{itemize}
\item A value of 3, termed mesokurtic, indicates the perfect conformance to the normal distribution.
\item A larger Kurtosis greater than 3, known as leptokurtic, exhibits a narrower peak.
\item A lower Kurtosis less than 3, referred to as platykurtic, corresponds to a wider peak and flatter tails.
\end{itemize}


Existing studies~\cite{lin_awq_2023,kim_squeezellm_2023} have revealed that
preserving outliers is crucial for achieving excellent quantization accuracy. The
presence of outliers in a particular layer can be identified using layer-wise
Kurtosis metrics make Kurtosis a valuable indicator for determining layers
that are challenging to quantify. Layers with the highest Kurtosis values can be
isolated using the outlier detection algorithm discussed in the following
section.

To measure Kurtosis metrics, the Pearson definition for each quantizable weight
matrix is employed to pre-calculate known models by leveraging the
\verb=scipy.stats= library~\cite{2020SciPy-NMeth}. The tool and instructions to
generate Kurtosis metrics are described in
Appendix~\ref{sec:kurt-metric-prep-tool}. For practical application in
the production environment, the Kurtosis metrics could be calculated on the fly
since the calculation is relatively lightweight.

\subsection{Outlier Detection Algorithm}
\label{sec:outlier-detection}

The outlier detection algorithm is designed to single out layers with extreme
sensitivity or Kurtosis values so that an additional bit budget can be allocated
to improve accuracy. Outliers are usually a small portion of the overall dataset.
The proposed outliers detection algorithm is capable of prioritizing top
outliers by rate of change so that only a limited surplus bit budget is
allocated. The layer-wise sensitivity or Kurtosis dataset is formulated in
Equation~\ref{eq:sensitivity-set}.
\begin{equation}
  S=\left\{s_1,s_2,\cdots,s_n\right\}
  \label{eq:sensitivity-set}
\end{equation}
The difference between two adjacent layers, denoted as the set $D$, is defined
in Equation~\ref{eq:sensi-diff-set} to filter the sensitive data points.
 \begin{equation}
  D = \left\{s_2-s_1, s_3-s_2,\cdots,s_n-s_{n-1}\right\}
  \label{eq:sensi-diff-set}
 \end{equation}

For datasets with an approximately ascending pattern, an alternative difference
set, as denoted in Equation~\ref{eq:sensi-ratio-set}, is defined to use
division instead of subtraction, with the advantage of ignoring data
points restoring to normal range. This is beneficial in reducing false alarms and
economize bit budget.

 \begin{equation}
  D = \left\{\frac{s_2}{s_1}, \frac{s_3}{s_2},\cdots,\frac{s_n}{s_{n-1}}\right\}
  \label{eq:sensi-ratio-set}
 \end{equation}

The set $D$ is assumed to follow the normal distribution. Therefore, the z-score can
be leveraged to isolate the outliers. Constrained by memory, only top-$m$
outliers are considered. Equation~\ref{eq:outlier-filter} defines the rule
to identify top-$m$ sensitive layers:
\begin{equation}
  \begin{aligned}
    Top_m(D') &= \left\{d \in D' \mid Rank(d, D') \le m \right\} \\
    D' &= \left\{d \in D \mid \frac{|d - \mu|}{\sigma} > 3 \right\} \\
    \sigma &= \frac{1}{n-1} \cdot \sqrt{\sum_i^{n}(d_i-\mu)^2} \\
    \mu &= \frac{1}{n} \cdot \sum_i^{n}d_i \\
  \end{aligned}
  \label{eq:outlier-filter}
\end{equation}
where $Rank(d,D')$ gives the position of $d$ when $D'$ is sorted in descending
order. To avoid the influence of extreme values, the mean and standard
deviation is calculated using the trimmed approach~\cite{Wilcox2010}, where
the 5\% smallest and largest data points are discarded. The actual
implementation leverages the \verb=scipy.stats= library. Finally, the sensitive
layer can be identified by Equation~\ref{eq:sensi-layer-set}, which adds 1 to
the indices returned by Equation~\ref{eq:outlier-filter} since $|D| = |S| - 1 =
n-1$.

 \begin{equation}
  \begin{aligned}
    I' &= \{i+1 \mid i \in ord(x, Top_m(D'))\, \forall x \in Top_m(D')\} \\
    ord(x,S) &= \{i \mid x_i = x, x_i \in S, i \in \{1,2,\cdots, n\}\} \\
  \end{aligned}
  \label{eq:sensi-layer-set}
 \end{equation}


\begin{algorithm}
\caption{The outlier detection algorithm}
\label{alg:outlier-detection}
\begin{algorithmic}[1]
\Input
\Desc{$S$}{Array of sensistivity scores or kurtosis metrics}
\Desc{$m$}{Number of top outliers to return}
\Desc{$t$}{Method to construct the difference set}
\EndInput
\Output
\Desc{$I'$}{Array of outlier indices}
\EndOutput
\State \textbf{Require} $S = \{s_1,s_2,\cdots,s_n\}$
\State \textbf{Ensure} $m \geq 1$
\State \textbf{Ensure} $t == $`subtract' or $t == $`divide'
\If{$t$ == `subtract'}
    \State $D \gets \left\{s_2-s_1, s_3-s_2,\cdots,s_n-s_{n-1}\right\}$
\Else
    \State $D \gets \left\{\frac{s_2}{s_1}, \frac{s_3}{s_2},\cdots,\frac{s_n}{s_{n-1}}\right\}$
    \Comment{suppress data points restore to normal range}
\EndIf
\State $\mu \gets \frac{1}{n} \cdot \sum_i^{n}d_i$
\State $\sigma \gets \frac{1}{n-1} \cdot \sqrt{\sum_i^{n}(d_i-\mu)^2}$
\State $i, n \gets 0, |D|$
\State $D' \gets \{\}$
\While{$i \leq n$}
\State $z_i \gets \frac{|d_i-\mu|}{\sigma}$
\If{$z_i > 3$}
\State $D' \gets D' \cup \{(d_i,ord(z_i)\}$
\EndIf
\EndWhile
\State $D'_m \gets sort(D')[:m]$
\State $I' \gets \{\}$
\State $i, n \gets 0, |D'_m|$
\While{$i \leq n$}
\State $I' \gets I' \cup \{d'_{m_{_i}}[1]+1\}$
\EndWhile
\State \Return $I'$
\end{algorithmic}
\end{algorithm}

Algorithm~\ref{alg:outlier-detection} presents the pseudo-code to locate the
outliers, given an array of sensitivity scores or Kurtosis metrics.


\subsection{SensiBoost and KurtBoost}

This section describes SensiBoost and KurtBoost, the two methods leveraging
activation sensitivity and Kurtosis metrics to enhance quantization accuracy
with a minimal increment in the bit budget. The new approaches are implemented by
identifying the sensitive layers using the outlier detection algorithm
explained in the previous Section~\ref{sec:outlier-detection}.

The key steps of the SensiBoost and KurBoost are as follows:
\begin{enumerate}
  \item Load the pre-calculated sensitivity scores or Kurtosis metrics for the model being quantized.
  \item Identify layers for additional allocation using the outlier detection algorithm according to the top-$m$ setting.
  \item Allocate normal budget to non-sensitive layers and assign additional budget to sensitive layers according to the boost stop setting.
  \item Apply quantization using the underlying quantization method.
\end{enumerate}


\begin{table}[h]
\centering
\caption[HQQ bit budgets]{HQQ bit budgets}
  \label{tab:hqq-bit-budgets}
\setlength{\tabcolsep}{11.0pt}
\resizebox{0.85\textwidth}{!}{
\begin{tabular}{r|c|c|c|c|c|r|c|c|c|c|c}
\toprule
\textbf{stop} & \textbf{budget} & $b_1$ & $g_1$ & $b_2$  & $g_2$  &
\textbf{stop} & \textbf{budget} & $b_1$ & $g_1$ & $b_2$  & $g_2$  \\
\midrule
0 & 2.13 & 2 & 128 & 8 & 128 & +1 & 2.25 & 2 & 64 & 8 & 128 \\
+2 & 2.51 & 2 & 32 & 8 & 128 & +3 & 3.13 & 3 & 128 & 8 & 128 \\
+4 & 3.25 & 3 & 64 & 8 & 128 & +5 & 3.51 & 3 & 32 & 8 & 128 \\
+6 & 4.13 & 4 & 128 & 8 & 128 & +7 & 4.25 & 4 & 64 & 8 & 128 \\
+8 & 4.51 & 4 & 32 & 8 & 128 & +9 & 8.13 & 8 & 128 & 8 & 128 \\
+10 & 8.25 & 8 & 64 & 8 & 128 & +11 & 8.51 & 8 & 32 & 8 & 128 \\
\bottomrule
\end{tabular}
}
\end{table}

To apply additional memory allocation, the amount of surplus budget for the
sensitive layers can be specified by the number of \textbf{boost stops}. When
boost stops go beyond the maximum bit budget of the underlying quantization
method, the maximum bit budget takes effect. Table~\ref{tab:hqq-bit-budgets}
presents the 12-stop bit budgets on top of HQQ. For instance, when the base bit
budget is 4.13, a setting of 2-stop will quantize the sensitive layers with a
bit budget of 4.51. However, when the base bit budget is 8.25, a 2-stop
increment request only results in 1 stop, i.e., a bit budget of 8.51.

The number of layers targeted for extra allocation can be restricted based
on the descending rank of sensitivity scores or Kurtosis metrics. The resulting
layers, referred to as top-$m$ layers, enable further control over the
allocation of a limited extra memory budget. Depending on the number of outliers
identified, the actual layers eligible for additional allocation might be fewer
than the specified value $m$. These layers may also vary across modules. No extra memory is assigned to modules without evident outliers. Lastly,
when $m$ is set to 0, all layers identified by the outlier detection algorithm
are considered for additional allocation.

\subsection{Experiments}

To assess the effectiveness of the proposed SensiBoost and KurtBoost methods,
models quantized using the two approaches were evaluated using the WikiText-2
and C4 datasets to measure the perplexity scores. For each proposed method,
various boost stop and top-$m$ configurations were benchmarked. Specifically,
these experiments involved benchmarking two boost stop settings (2 and 3) and
four top-$m$ values (1, 2, 3, and 0) across three Llama models under six base-bit budget configurations. Furthermore, ablation studies were included to
validate the efficacy of the proposed methods. The ablation tests randomly
select the layers from a set that explicitly excludes the layers identified by
SensiBoost or KurtBoost. To ensure a fair comparison,
the amount of extra memory and the layers are identical to those used in
SensiBoost or KurtBoost. The complete permutations of the test cases consist
of a total of 576 test cases.


\begin{table}[t]
\centering
\caption[Assessment matrix of various approaches]{Assessment matrix of various approaches}
\label{tab:method-eval-matrix}
\setlength{\tabcolsep}{12.0pt}
\resizebox{0.5\textwidth}{!}{
\begin{threeparttable}
\begin{tabular}{l|c|c|c|c|c|c}
\toprule
\textbf{Method} & SB & KB & SBAB & KBAB & HQQ & MXQ \\
\midrule
SB\tnote{1}   & - & X & X & - & X & X \\
KB\tnote{2}   & - & - & - & X & X & X \\
SBAB\tnote{3} & - & - & - & - & - & - \\
KBAB\tnote{4} & - & - & - & - & - & - \\
HQQ  & - & - & - & - & - & - \\
MXQ  & - & - & - & - & - & - \\
\bottomrule
\end{tabular}
\begin{tablenotes}
\item[1] {\scriptsize SB denotes the SensiBoost method.}
\item[2] {\scriptsize KB denotes the KurtBoost method.}
\item[3] {\scriptsize SBAB denotes the ablation test for SensiBoost method.}
\item[4] {\scriptsize KBAB denotes the ablation test for KurtBoost method.}
\end{tablenotes}
\end{threeparttable}
}
\end{table}

The comparisons of the different approaches were made among SensiBoost,
KurtBoost, corresponding ablation methods, HQQ, and MXQ, which are presented in
Table~\ref{tab:method-eval-matrix}. Win-tie-loss scores were used to
qualitatively analyze the proposed methods. These scores were aggregated from
the perplexity results benchmarked and paired based on
Table~\ref{tab:method-eval-matrix}. Specifically, all perplexity scores were
rounded to two decimal places. The perplexity of the primary method (SensiBoost
or KurtBoost) was then subtracted from the comparison method to
determine the win-tie-loss score. A negative difference awarded the primary
method 1 win, a difference of zero awarded 1 tie, and a positive difference
awarded 1 loss. Finally, the win-tie-loss scores were aggregated across six
quantization configurations, two stop settings, four top-$m$ settings, and two
evaluation datasets, providing a summarized win-tie-loss analysis for various
method pairs across the three Llama models.


\begin{figure*}[t!]
  \centering
\includegraphics[width=0.95\textwidth]{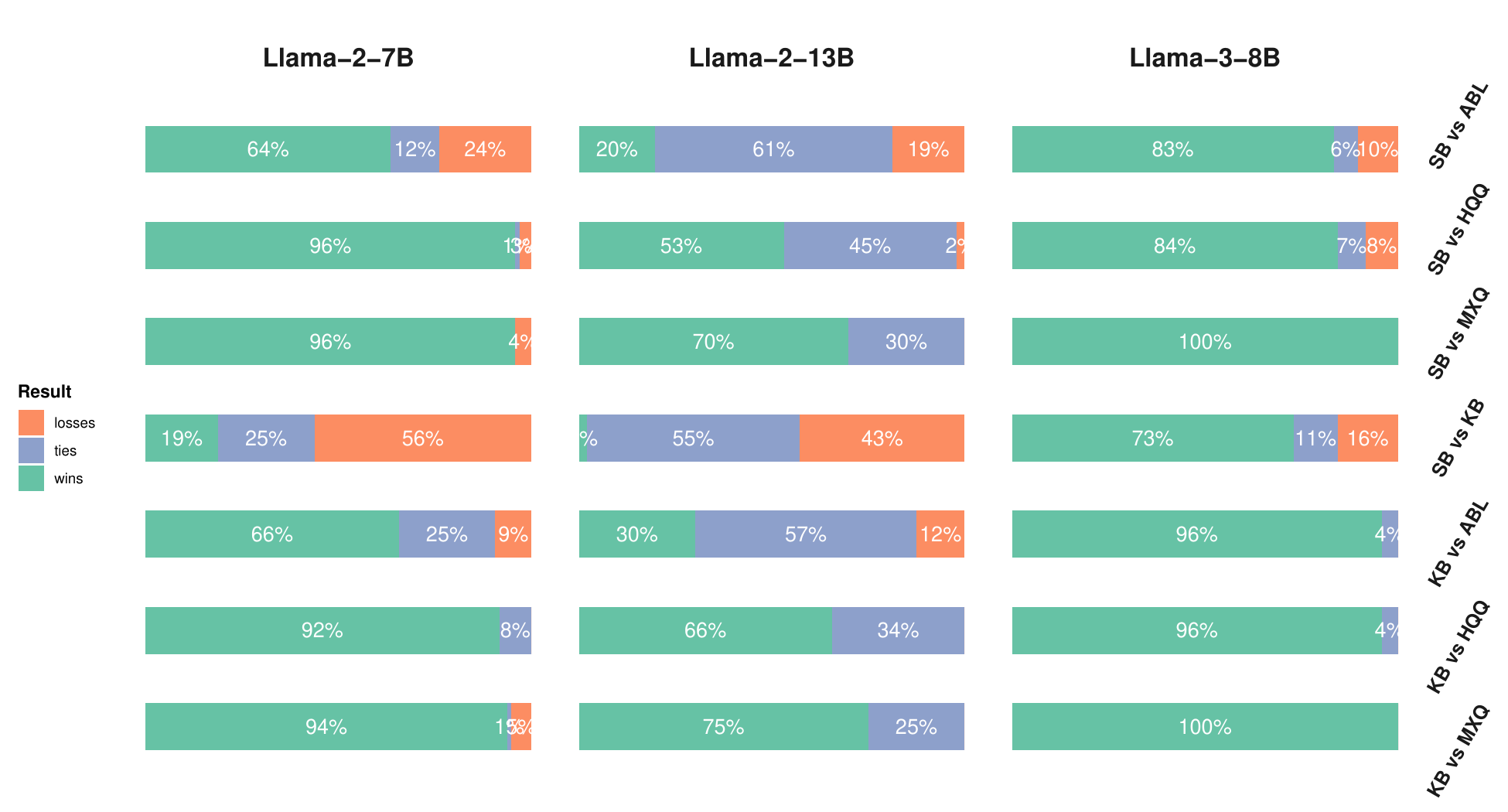}
\caption[Win-tie-loss diagram of SensiBoost/KurtBoost method]{
This figure illustrates the win-tie-loss performance of the SensiBoost (denoted
as "SB") and KurtBoost (denoted as "KB") methods compared to their ablation
test (labeled as "ABL") as well as the baseline methods HQQ and MXQ, across
three Llama models. As anticipated, SensiBoost and KurtBoost outperform the
baseline methods HQQ and MXQ due to the allocation of additional bit budgets.
However, their relatively low win rates (53\% against HQQ and 70\% against MXQ
in the case of SensiBoost, 66\% against HQQ and 75\% against MXQ for KurtBoost)
on the Llama-2-13B model suggest that achieving significant improvements in
larger models with a limited extra memory budget is challenging. SensiBoost
consistently outperforms its ablation test variant. However, its comparison
with the KurtBoost method reveals mixed outcomes: while SensiBoost
underperforms on the two Llama-2 models, it demonstrates considerable
advantages on the Llama-3-8B model. For optimal clarity, the figure is best
viewed in color and with zoom.
}
\label{fig:wtl-llama-comparison}
\end{figure*}


\begin{figure}[t!]
  \centering
\includegraphics[width=\textwidth]{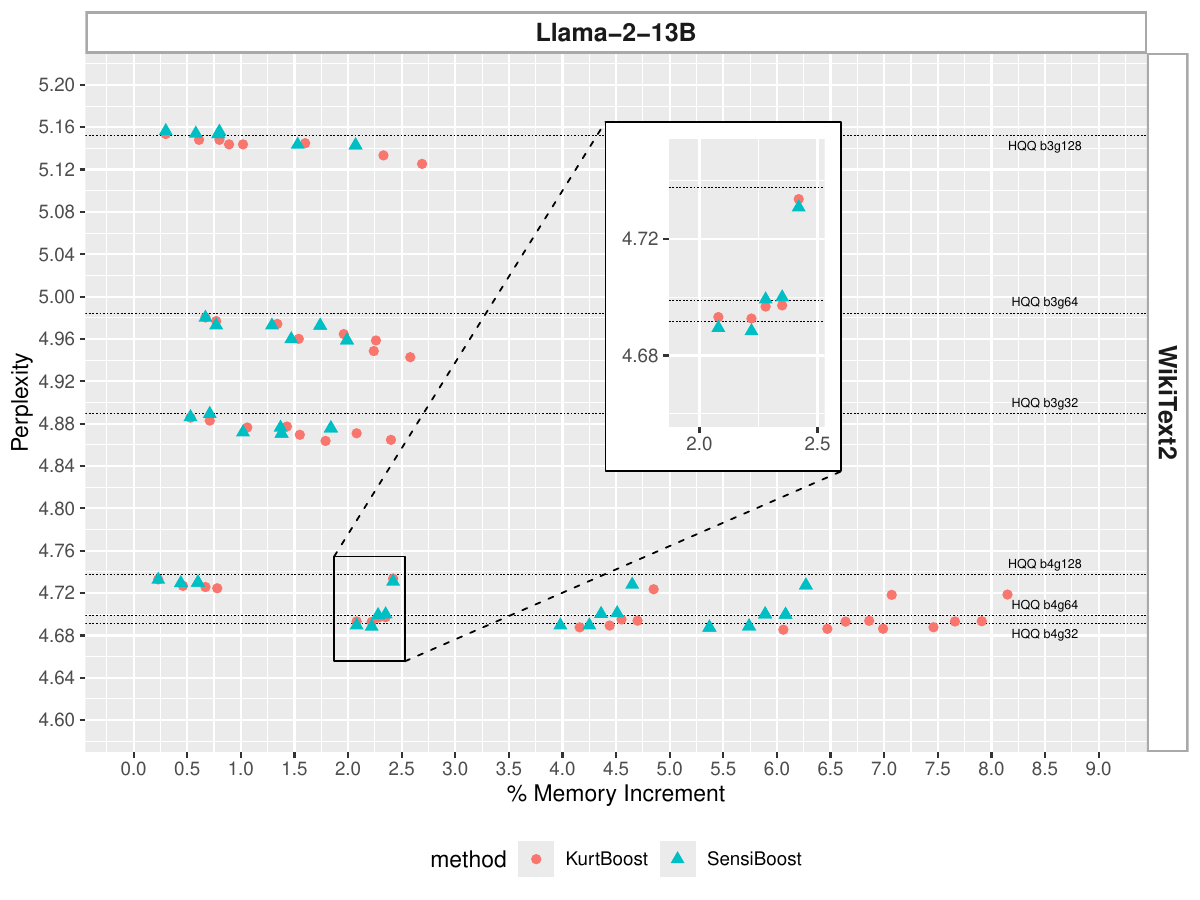}
\caption[SensiBoost vs KurtBoost Performance Comparison on WikiText2]{
This figure illustrates the perplexity performance of the SensiBoost and
KurtBoost approaches evaluated on the Llama-2-13B model using the WikiText2
dataset. The green triangles, representing the SensiBoost method, are
positioned closer to the y-axis, indicating that SensiBoost requires less
additional memory to achieve comparable performance to KurtBoost. Notably,
SensiBoost exhibits a slight advantage over KurtBoost, requiring approximately
2\% more bit budget to attain a near-minimal perplexity score, as emphasized in
the magnified sub-plot. For optimal interpretation, the figure is best viewed
in color and with zoom.
}
\label{fig:ppl-sk-Llama-2-13b-wikitext}
\end{figure}


\begin{figure*}[t]
  \centering
\includegraphics[width=0.95\textwidth]{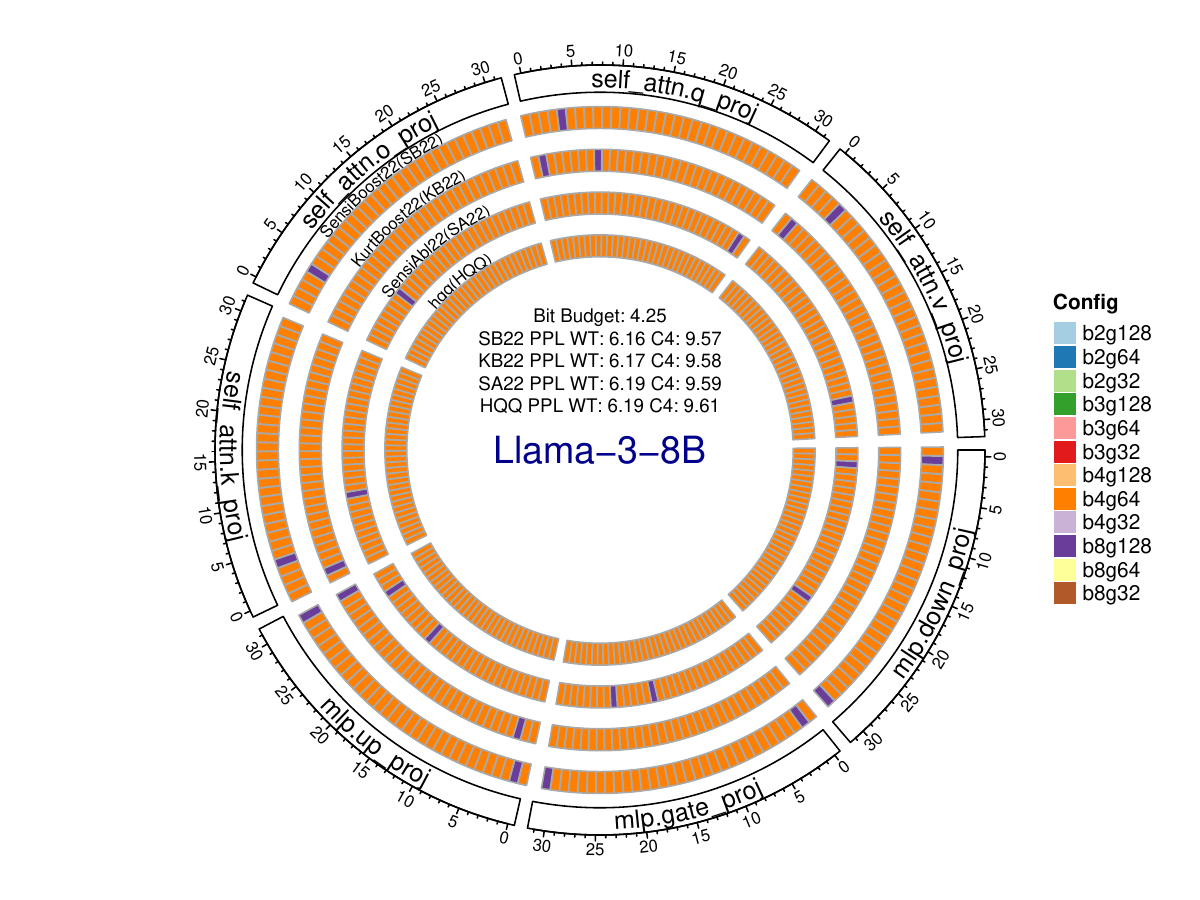}
\caption[Optimal SensiBoost Quantization Config Allocation Example]{
This figure presents a comparison of quantization configuration allocations as
determined by SensiBoost, KurtBoost, HQQ and MXQ for the Llama-3-8B model,
under a bit budget of 4.25. The colored rings represent the assigned
quantization configurations denoted as b2g128 through b8g32, while the text in
the center indicates the perplexity scores. As demonstrated by this figure the
SensiBoost with a boost stop value of 2 and top-$m$ 1, as denoted by SB22,
yields a perplexity score of 6.16 on WikiText2, 9.57 on C4, outperforming the
HQQ baseline, its ablation variant (SA22) and KurtBoost (KB22). For optimal
clarity, the figure is best viewed in color and with zoom.
}
\label{fig:llama-8b-cfg-allocation-sb22}
\end{figure*}


\begin{figure*}[t!]
  \centering
\includegraphics[width=0.95\textwidth]{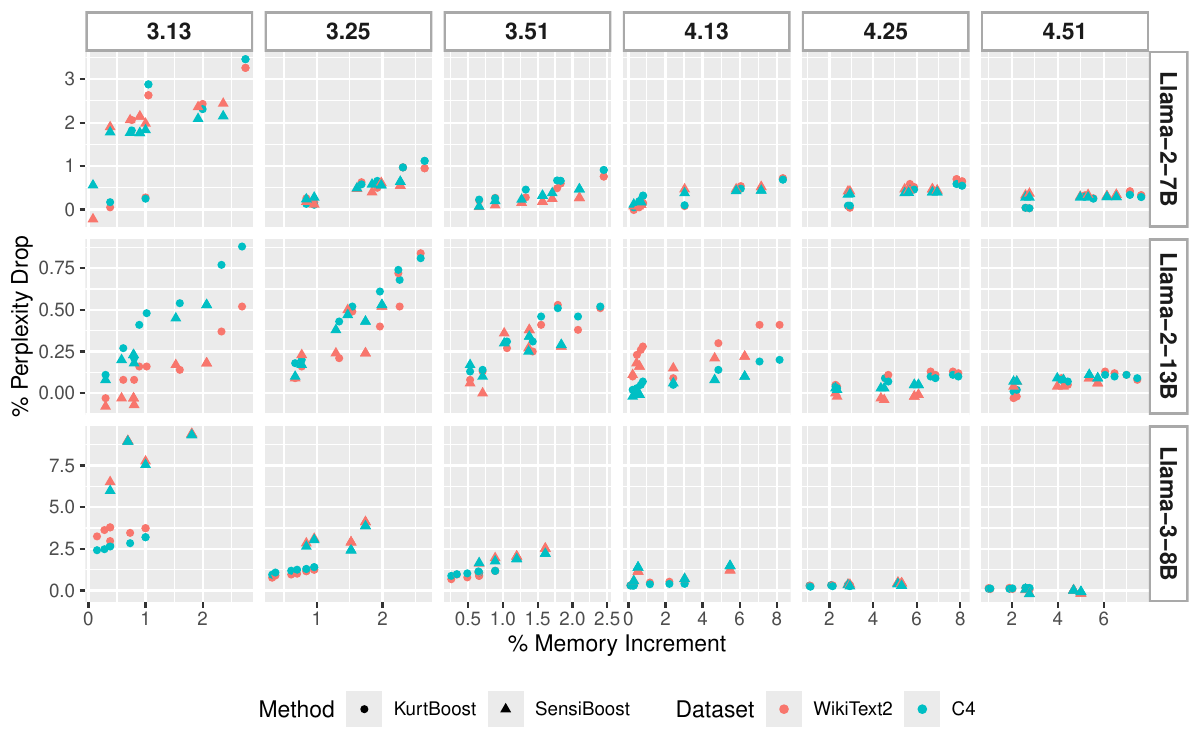}
\caption[Perplexity Drop vs Memory Budget Increment]{
This figure compares the SensiBoost and KurtBoost methods' performance on
quantization accuracy enhancement with additional budget. The X axis denotes
the percentage of additional budget assiged. The Y axis represents the
percentage of perplexity drop. The improvement is more pronounced around the
3-bit range on the two smaller models Llama-2-7B and Llama-3-8B respectively.
Notably, SensiBoost, denoted as triangles, exhibits more aggressive improvement
with perplexity drop up to 9\% on Llama-3-8B under a budget of 3.13. This
figure also demonstrates the challenge of achieving substantial quantization
accuracy elevation at higher bit budgets. For optimal clarity, the figure is
best viewed in color and with zoom. }
\label{fig:ppl-decr-extra-budget}
\end{figure*}

\subsection{Ablation Test}

Ablation studies were integrated into the experiments to validate the
effectiveness of the proposed SensiBoost and KurtBoost methods, ensuring they
outperform random choices. The ablation tests were carried out using random
selection and explicitly avoiding choosing layers that could be potentially
selected by either SensiBoost or KurtBoost.

Formally, given $I'$ as defined in Equation~\ref{eq:sensi-layer-set}, $I'_{s}$ is
the sensitive layers identified by SensiBoost, $I'_{k}$ denotes the ones for
KurtBoost, the corresponding ablation test layer choices $J_{s}$ and $J_{k}$
are defined by Equation~\ref{eq:ablation-choice}.
\begin{equation}
  \label{eq:ablation-choice}
  \begin{aligned}
    J_{s} &= \{h_1,h_2,\cdots,h_p \mid h_i \in \hat{I},h_i\neq h_j \forall i \neq j, p = |I'_{s}| \} \\
    J_{k} &= \{h_1,h_2,\cdots,h_q \mid h_i \in \hat{I},h_i\neq h_j \forall i \neq j, q = |I'_{k}| \} \\
    \hat{I} &= I \setminus (I'_{s} \cup I'_{k}) \\
    I &= \{1,2,\cdots,n \} \\
  \end{aligned}
\end{equation}
where $n$ is the number of layers in a particular large language model.

For example, suppose the set $\{2, 31\}$ represents the sensitive layers
discovered by the SensiBoost method, whereas the set $\{1, 30, 31\}$ is
identified by the KurtBoost approach, then the set $\{3, 28\}$ is a valid
choice for ablation test of SensiBoost under top-$m=3$. Likewise, the $\{4,
28, 29\}$ is a legitimate ablation test configuration for KurtBoost. However,
the set $\{2, 28, 29\}$ is invalid for ablation test of KurtBoost, since it
contains the layer 2 which could potentially enhance quantization accuracy
since it is considered as a sensitive layer by the SensiBoost method.

This paper includes two sets of ablation tests designed to validate the
efficacy of the SensiBoost and KurtBoost approaches. These tests were conducted
under the same configurations as their non-ablation counterparts. Specifically,
the configurations consist of two boost stop values and four top-$m$ settings,
evaluated across three Llama models under six base bit budget configurations.


\subsection{Results and Analysis}

The overall results are presented in this section to qualitatively assess the
effectiveness of SensiBoost and KurtBoost by leveraging win-tie-loss
comparison. The win-tie-loss diagram, presented in
Figure~\ref{fig:wtl-llama-comparison}, includes the comparisons between the
proposed methods (indicated by the row labels) and their ablation variants as
well as baselines such as HQQ and MXQ, across three Llama models (denoted by
the column labels).

As anticipated, both SensiBoost and KurtBoost outperform the baseline methods
HQQ and MXQ due to the allocation of additional bit budgets. However,
SensiBoost's relatively low win rates (53\% against HQQ and 70\% against MXQ)
on the Llama-2-13B model suggests that achieving significant improvements in
larger models with a limited extra memory budget is challenging. KurtBoost
performs slightly better than SensiBoost on the Llama-2-13B, achieving 66\% win
rate against HQQ and 75\% against MXQ.

In the context of ablation testing, both methods generally outperform their
ablation variants. However, the 20\% win rate and 61\% tie rate on the
Llama-2-13B model suggests that SensiBoost struggles to surpass its ablation
counterpart when applied to larger models. In contrast, KurtBoost consistently
demonstrates a slight advantage over SensiBoost, achieving higher win rates and
lower loss rates across all three models.

\subsection{SensiBoost and KurtBoost Comparison}



The previous section provides an overall comparison of the SensiBoost and
KurtBoost approaches. A detailed and direct comparison of the two methods is
presented in this section to reveal the relative advantages and disadvantages
of the two methods under various scenarios.

As indicated by the win-tie-loss result in
Figure~\ref{fig:wtl-llama-comparison}, SensiBoost tends to be less performant
than KurtBoost. However, a deeper examination reveals that SensiBoost requires
less additional memory to achieve comparable performance to KurtBoost on the
Llama2-7B and Llama-2-13B models. As demonstrated in
Figure~\ref{fig:ppl-sk-Llama-2-13b-wikitext}, SensiBoost exhibits a slight
advantage over KurtBoost in identifying optimal quantization configuration
where it requires approximately 2\% more bit budget to attain a near-minimal
perplexity score, as highlighted in the magnified subplot. This phenomenon
replicates to the C4 dataset and the Llama2-7B model.

On the other hand, however, the situation is reversed on the Llama-3-8B model,
where KurtBoost is more effective in discovering optimal quantization
configuration, which is both memory-efficient and yields better accuracy. The
Kurtosis metrics are generated individually for distinct modules in each layer.
In contrast, the sensitivity scores are measured by following the
neural-network computation sequence, where some modules share the same
sensitivity score as they are not standalone computation units. Therefore, the
KurtBoost approach may yield more nuanced quantization configurations that are
more memory-efficient, as demonstrated in
Figure~\ref{fig:llama-8b-cfg-allocation-sb22}, where the
\verb=self_attn.o_proj=, \verb=mlp.down_proj=, and \verb=mlp.gate_proj= modules
are not assigned extra budget.


In conclusion, both SensiBoost and KurtBoost demonstrate advantages over the
baselines and their ablation variants, indicating the effectiveness of the two
approaches. Both methods enable model accuracy enhancement by using
approximately 2\% additional memory budget. Specifically, the improvement is
more pronounced in the 3-bit range with perplexity drop up to 9\% on Llama-3-8B
archived by the SensiBoost as illustrated in
Figure\ref{fig:ppl-decr-extra-budget}. This figure also demonstrates the
challenge of achieving substantial quantization accuracy improvements at higher
bit budgets, as evidenced by the notably flat pattern in the sub-plots for 4.25
and 4.51 configuration.

\section{Conclusion and Future Work}



This paper presents a novel approach to improving quantization accuracy in LLMs by incorporating layer-wise sensitivity analysis. The study empirically explores the impact of quantization errors across multiple transformer-based LLM families, revealing that sensitivity patterns remain consistent within a model family and its fine-tuned variants. This observation provides valuable insights into the structural characteristics of large-scale neural networks and highlights the need for adaptive quantization strategies.

In the proposed layer-sensitive approach, an outlier detection algorithm is introduced to identify layers that are particularly sensitive to quantization errors. By utilizing activation sensitivity scores and weight distribution Kurtosis metrics, the proposed approach effectively detects layers that require differentiated memory allocation. Building upon these insights, the SensiBoost and KurtBoost methods are developed to selectively allocate additional memory to the most sensitive layers while maintaining an overall memory budget. Experimental results demonstrate that these methods achieve superior quantization accuracy, outperforming the state-of-the-art HQQ approach. Specifically, the proposed techniques lead to a reduction in perplexity of up to 9\% while increasing the memory budget by only 2\%, striking a balance between efficiency and performance.

The findings of this work suggest that leveraging layer-wise sensitivity features, such as activation sensitivity and Kurtosis, enables more effective quantization strategies with minimal additional computational cost. By integrating these methods into existing quantization frameworks, it becomes possible to enhance the efficiency of LLM deployment without sacrificing model accuracy.

Future research could extend this sensitivity analysis to a broader range of transformer architectures and explore more sophisticated approaches to dynamically adjusting quantization configurations based on computational constraints. As the demand for efficient LLM deployment continues to grow, sensitivity-aware quantization techniques will play a crucial role in optimizing model performance while maintaining practical resource requirements.


\bibliographystyle{unsrt}
\bibliography{references}

\appendix
\section{The lm-quant-toolkit overview}

The \verb=lm-quant-toolkit= is a suite of tools to facilitate large neural
network quantization research. It includes a quantization harness tool to drive
quantization experiments on large language models and vision models, to collect
and summarize experiment data for further analysis. It also includes tool to
prepare experiment meta data and visualization tools to interpret experiment
results. Specifically, \verb=lm-quant-toolkit= consists of:

\begin{itemize}
  \item LLM quantization harness tool
  \item Kurtosis Metrics Measuring Tool
  \item Sensitivity Score Measuring Tool
  \item Calibration Dataset Generation Tool
\end{itemize}

Most tools are implemented in Python and are extensively tested under the
Python 3.11.9. The visualization tools are implemented in R. The usage of
these tools is elaborated in the following sections.

The Python tools depend on Python libraries such as transformers, datasets,
numpy, PyTorch, among others. A few Python libraries are patched to support the
proposed quantization methods. Specifically, required patched dependencies
include AutoGPTQ (for CUDA 12.5 compatibility), HQQ (support
SensiBoost/KurtBoost extension), lm\_eval (for end-to-end LLM performance
evaluation), clip\_benchmark (for vision model evaluation). These dependencies
are installed automatically as part of setup process.

The visualization tools facilitate visualizing the experiment results, the
weight distribution, and generating insights of the latent features to quantize
LLMs more efficiently. Most visualization tools are implemented in R and
leverage the open-source plot libraries such as ggplot2~\cite{wickham_2016},
circlize~\cite{zuguang_2014}, ggbreak~\cite{shuangbin_2021},
and ggmagnify~\cite{hugh_2024}.

To setup the test harness and visualization tools, follow the instructions
on \href{https://github.com/schnell18/lm-quant-toolkit.git}{https://github.com/schnell18/lm-quant-toolkit.git}.

%

\subsection{Kurtosis Metrics Measuring Tool}
\label{sec:kurt-metric-prep-tool}

This tool calculates the Kurtosis metrics of weight matrices layer-by-layer
inside a particular large language model. The Kurtosis metrcis are crucial to
identify sensitive layers to improve the accuracy of quantization. This tool
accepts a list of Hugging Face-compliant model identifiers. The output of this
tool is a series of .csv files under specified directory. Each file contains
the Kurtosis metrics for corresponding models.

The tool is implemented in Python and provides a convenient CLI interface to
enable shell scripting. It is included in the \verb=dump.py= file under the
\verb=src= folder in the \verb=lm-quant-toolkit= project.

\subsection{Sensitivity Score Measuring Tool}
\label{sec:sensi-measuring-tool}

This tool calculates the sensitivity scores of each layer of a particular large
language model. The sensitivity scores are crucial to identify sensitive layers
to improve the accuracy of quantization. This tool accepts a list of Hugging
Face-compliant model identifiers. The output of this tool is a series of .csv
files, each containing the sensitivity score for corresponding model. These
files are crucial inputs to guide the SensiBoost and Sensitivity-based MiLP.

The tool is implemented in Python and provides a convenient CLI interface to
enable shell scripting. It is compatible with any transformer-based LLMs with
an implementation of the popular Hugging Face transformers library. It is
located separately in the \verb=dump.py= file under the \verb=src= folder in
the \verb=lm-quant-toolkit= project, which helps to reduce unnecessary
dependencies. A typical usage is demonstrated in the code snippet as follows:

The code snippet demonstrates how to calculate the sensitivity scores for a
series of Qwen2.5 models using 4 calibration datasets under 12 bit budgets.

\subsection{Calibration Dataset Generation Tool}
\label{sec:bos-gen-tool}

This tool generates a small synthesized dataset named Branch of Science
(denoted as BoS, published on Hugging Face), which includes a few hundred of
textual defintions for science, art and business topics such as Mathematics,
Physics, Chemstry, Law, Music and Journalism, among others. The dataset is
intended to validate whether the sensitivity property generalizes to diverse
datasets.

The tool generates an initial dataset in .csv format which requires further
processing. The output of this tool is random due to the generative nature of
LLM. This tool requires a Llama-2-7B model being served with an OpenAI
compatible RESTful API endpoint. User can either use a hosted API endpoint or
deploy a local instance by following the instruction at the end of this
section.




\end{document}